\documentclass{article}
\usepackage[tikz]{KLMM/klmm}
\usepackage{tabularx}
\usepackage{xurl}
\newcolumntype{Y}{>{\raggedright\arraybackslash}X}
\usepackage{enumitem}
\usepackage{makecell}
\usepackage{pifont}
\usepackage{subcaption}
\usepackage{pgfplots}
\usepackage{booktabs}
\usepackage{multirow}
\usepackage{threeparttable}
\usepackage{xcolor}
\usepackage{listings}
\usepackage[most]{tcolorbox}
\tcbuselibrary{skins,breakable}

\tcbuselibrary{listings, breakable, skins}

\newtcblisting{contentbox}[1][Model Input (JSON)]{
    listing only,
    breakable,
    enhanced,
    title={#1},
    fonttitle=\bfseries\small,
    coltitle=black,
    colbacktitle=gray!12,
    colback=gray!3,
    colframe=gray!45,
    boxrule=0.4pt,
    arc=1mm,
    left=2mm,
    right=2mm,
    top=1.5mm,
    bottom=1.5mm,
    listing options={
        basicstyle=\ttfamily\scriptsize,
        columns=fullflexible,
        breaklines=true,
        breakatwhitespace=false,
        keepspaces=true,
        showstringspaces=false,
        numbers=none,
        tabsize=2
    }
}
\newtcolorbox{promptbox}[1]{
    enhanced,
    breakable,
    width=\linewidth,
    colback=black!2,
    colframe=black!45,
    colbacktitle=black!8,
    coltitle=black,
    title={#1},
    fonttitle=\small\bfseries,
    fontupper=\small,
    boxrule=0.5pt,
    arc=2pt,
    left=8pt,
    right=8pt,
    top=6pt,
    bottom=6pt,
    before skip=10pt,
    after skip=10pt
}

\pgfplotsset{compat=1.18}

\lstdefinestyle{promptjson}{
    language={},
    basicstyle=\ttfamily\footnotesize,
    columns=fullflexible,
    breaklines=true,
    breakatwhitespace=false,
    keepspaces=true,
    showstringspaces=false,
    numbers=none,
    frame=none,
    aboveskip=6pt,
    belowskip=0pt
}

\usepackage{float}  

\floatstyle{ruled}
\newfloat{listing}{tb}{lst}
\floatname{listing}{Listing}
\floatstyle{plain}
\newfloat{reviewcase}{tbp}{lor}
\floatname{reviewcase}{Case}
\DeclareCaptionStyle{ruled}{labelfont=normalfont,labelsep=colon,strut=off} 
\usepackage{amsfonts}
\usepackage{amsopn}
\definecolor{keywordcolor}{rgb}{0.7, 0.1, 0.1}
\definecolor{tacticcolor}{rgb}{0.0, 0.1, 0.6}
\definecolor{commentcolor}{rgb}{0.4, 0.4, 0.4}
\definecolor{symbolcolor}{rgb}{0.0, 0.1, 0.6}
\definecolor{sortcolor}{rgb}{0.1, 0.5, 0.1}
\definecolor{stringcolor}{rgb}{0.5, 0.3, 0.0}

\lstnewenvironment{leancode}[1][]{%
  \lstset{language=lean}%
  \lstset{mathescape=false,extendedchars=true}%
  #1%
}{}

\title{LeanReviewBench: Semantic Alignment Review in Mathematical Lean Projects}
\author{Ruichen~Qiu\institution{State Key Laboratory of Mathematical Sciences, Academy of Mathematics and Systems Science, Chinese Academy of Sciences, and University of Chinese Academy of Sciences. }\equalcontrib
    \and Hao~Shen\instref{1}\equalcontrib
    \and Lihong~Zhi\instref{1}
    \and Xiao-Shan~Gao\instref{1}
}

\begin{document}
\maketitle

\begin{abstract}
Lean verifies that a proof establishes its formal proposition, but whether that proposition preserves the mathematical content of an informal source requires separate review. In project-scale formalizations, this correspondence can depend on definitions and constraints distributed across many declarations. We introduce LeanReviewBench, a benchmark for project-level semantic alignment review containing 197 versioned inputs collected from 18 real Lean projects, with 87 aligned and 110 misaligned reference judgments. Each input binds an informal mathematical target to selected formal declarations and an extracted dependency context. Construction combines development-history discovery, independent target binding, reconstruction under a common Lean environment, and Astra annotation with human review of uncertain cases. The collection spans diverse mathematical domains and includes 69 historical pairs. We evaluate six language models with Direct and ReForm prompts on all 197 inputs against final human-reviewed reference labels. GPT-6.1 Sol with Direct prompting achieves the highest reported accuracy of 0.827 and misalignment-detection F1 of 0.844; the effects of prompting vary across models. LeanReviewBench provides a testbed for reviewing semantic correspondence in existing mathematical formalization projects.
LeanReviewBench is available at \url{https://github.com/MechMath/LeanReviewBench}.
\end{abstract}

\section{Introduction}
\label{sec:introduction}

AI systems are increasingly producing results on major open mathematical problems.
OpenAI has reported breakthroughs in areas including sphere packing and coding theory, followed by a broader release of mathematical manuscripts with accompanying Lean formalizations~\citep{openai2026advances,openai2026math}.
These developments make reliable verification of AI-generated mathematics increasingly important.
Lean provides machine checking of formal proofs, but the reliability of the resulting mathematical claims depends on both proof correctness and faithful specification: Lean checks that a proof establishes its formal proposition, while correspondence with the manuscript requires separate review.
For example, the HyperbolicObstruction manuscript in OpenAI's collection requires a two-dimensional simplicial complex, but the corresponding formalization omits this requirement, as its official scope documentation explicitly acknowledges~\citep{openai2026hyperbolic_scope}.
This example highlights the need to assess whether the formal statement preserves the mathematical content of its informal source.


In large Lean projects, alignment becomes more difficult, as identifying such discrepancies requires interpreting the definitions behind the selected declarations.
A theorem can use project-defined predicates, structures, and quantitative expressions whose meanings are distributed across files.
A condition absent from the displayed theorem type may be encoded in a structure field; conversely, a predicate with an appropriate name may omit a substantive requirement.
For the Brownian Motion project's main theorem, Lean Atlas illustrates a dependency cone of 227 nodes, from which Lean Compass selects 14 nodes for semantic verification~\citep{leanatlas}.
Semantic review therefore requires access to relevant mathematical dependencies, while much of the surrounding proof machinery can be excluded.

These observations motivate \textbf{LeanReviewBench}, a benchmark for semantic alignment review in existing Lean projects.
We inspect commit histories and associated development records to discover candidate discrepancies and recover the relevant formal snapshots.
Each selected snapshot is bound to an independently sourced informal target and supplied with extracted project dependencies, retaining mathematical definitions while excluding proof bodies.
GPT-6 Astra in Codex provides initial alignment annotations, with uncertain cases referred to a human annotator to determine whether the selected declarations preserve the target's mathematical content, allowing equivalent reformulations while treating strict strengthenings and weakenings as misalignment.
This construction brings together cases discovered in actual formalization work, project interpretation context, and traceable development history without using historical changes to predetermine alignment labels.

LeanReviewBench contains 197 versioned inputs from 18 contemporary Lean projects, covering diverse areas including analysis, number theory, geometry, probability, and applied modeling.
The formal inputs are reconstructed under a common Lean~4.34.1 environment, with informal scope and formal revisions fixed for reproducible evaluation.
Initial annotations are produced by GPT-6 Astra in Codex, with uncertain cases referred to a human annotator, yielding 87 aligned and 110 misaligned reference judgments.
The collection also includes 69 historical pairs linking 138 inputs; each version is assessed against its own target, so these pairs can capture both formal revisions and changes in mathematical scope.

We evaluate six language models using a minimal Direct prompt and the original ReForm prompt.
On all 197 inputs, GPT-6.1 Sol with Direct prompting obtains the highest reported accuracy, 0.827, and misalignment-detection F1, 0.844.
The effects of prompting vary across models.
These results provide baselines for semantic review with supplied dependency context.

Our contributions are:
\begin{itemize}[itemsep=2pt, topsep=0pt, parsep=0pt, leftmargin=*]
\item LeanReviewBench, an annotated collection of 197 versioned semantic alignment inputs from 18 contemporary Lean projects, including 69 historical pairs.
\item A construction pipeline that binds independent informal targets to fixed formal revisions and extracts their interpretation dependencies under a common Lean environment.
\item Baseline evaluations of six language models under two prompting strategies against final human-reviewed references.
\end{itemize}

\section{Related Work}
\label{sec:related}

\textbf{Informal–formal Semantic alignment.}
Informal--formal semantic alignment concerns whether formal statements preserve the mathematical content of their informal sources.
Methods for evaluating informal--formal semantic alignment include comparison with formal references, including structure-aware equivalence, direct assessment against informal sources, and backtranslation-based checks~\citep{beq,yu2026structure,formalalign,formalrx,claimcheck}.
Related proof-evaluation work uses faithfulness checks to gate formal evidence~\citep{wang2026faithsieve} or assesses correspondence between informal and formal proof processes~\citep{anonymous2026beyond}.
Benchmarks compared in Table~\ref{tab:alignment-benchmarks} use annotated informal--formal pairs to test whether these methods reliably distinguish faithful formalizations from semantic discrepancies.
FormalAlign~\citep{formalalign} and FormalRx-Test~\citep{formalrx} construct errors to test statement-level semantic assessment.
Additionally, ProofNetVerif~\citep{proofnetverif}, CriticLeanBench~\citep{criticlean}, ConsistencyCheck~\citep{reform}, MA-Align~\citep{mathatlas}, and DriftBench~\citep{driftbench} include errors discovered in existing formalizations, extending evaluation beyond constructed perturbations.
Meanwhile, FaithAlign~\citep{faithalign} supplies interpretation context and formal dependencies for phrase-level correspondence judgments.
Compared with above benchmarks, our LeanReviewBench combines project interpretation context, discovered errors, and development-history-based construction under Lean~4.34.1.


\begin{table}[t]
\centering
\footnotesize
\caption{Comparison of informal--formal semantic alignment benchmarks. Context indicates supplied project interpretation context; Discovered indicates errors identified in existing formalizations; History indicates collection using development records; Public indicates availability of inputs and reference labels. \ding{115} indicates unresolved error provenance in Discovered and availability of inputs only in Public. N/S indicates an unspecified Lean version.}
\label{tab:alignment-benchmarks}
\begin{tabular*}{\linewidth}{@{\hspace{10pt}\extracolsep{\fill}} l c c c c c @{\hspace{10pt}}}
\toprule
Benchmark & Lean & Context & Discovered & History & Public \\
\midrule
FormalAlign~\citep{formalalign} & 4.7.0-rc2 & \ding{55} & \ding{55} & \ding{55} & \ding{51} \\
ProofNetVerif~\citep{proofnetverif} & 4.8.0 & \ding{55} & \ding{51} & \ding{55} & \ding{51} \\
CriticLeanBench~\citep{criticlean} & N/S & \ding{55} & \ding{51} & \ding{55} & \ding{51} \\
ConsistencyCheck~\citep{reform} & N/S & \ding{55} & \ding{51} & \ding{55} & \ding{51} \\
MA-Align~\citep{mathatlas} & 4.24.0 & \ding{55} & \ding{51} & \ding{55} & \ding{51} \\
FormalRx-Test~\citep{formalrx} & 4.24.0 & \ding{55} & \ding{55} & \ding{55} & \ding{115} \\
DriftBench~\citep{driftbench} & N/S & \ding{55} & \ding{51} & \ding{55} & \ding{55} \\
FaithAlign~\citep{faithalign} & N/S & \ding{51} & \ding{115} & \ding{55} & \ding{55} \\
\midrule
\textbf{LeanReviewBench (ours)} & \textbf{4.34.1} & \ding{51} & \ding{51} & \ding{51} & \ding{51} \\
\bottomrule
\end{tabular*}
\end{table}

\textbf{Project-level formalization.}
Project-level formalizations rely on interdependent definitions, lemmas, and theorems.
Benchmarks extend formalization beyond isolated statements: RLM25 collects informal–formal pairs from blueprint-based Lean projects~\citep{proofnetverif}, LCS-Bench evaluates the formalization of an interdependent textbook theory~\citep{lcsbench}, and MathAtlas organizes mathematical content and its textbook dependencies~\citep{mathatlas}.
Project-level formalization workflows include Aria's dependency-driven decomposition and definition grounding~\citep{aria} and Lea's support for mathematician-controlled formalization~\citep{lea}.
For navigating and selecting relevant context, Lean Atlas visualizes project dependencies and provides the Lean Compass dependency-pruning procedure~\citep{leanatlas}.
Following these works, LeanReviewBench supplies project dependencies as an interpretation context for reviewing selected formal declarations.

\textbf{Audits of existing formalizations.}
MiniF2F-Lean Revisited and Ground False examine informal--formal discrepancies in existing benchmarks, with the latter releasing repaired material as ProofNet-Verified~\citep{minif2frevisited,groundfalse}.
At the project level, the numerical-analysis audit examines semantic correctness and library reuse~\citep{numericalaudit}, while LeanTrail assesses paper-to-Lean correspondence and proof completeness to report formalization coverage~\citep{leantrail}.
These studies motivate benchmarks for testing semantic review of existing formalizations beyond kernel acceptance.

\section{Benchmark}
\label{sec:benchmark}
LeanReviewBench organizes project-level semantic alignment review around mathematical claims and their versioned Lean realizations, retaining the context needed to interpret both.
This section defines the review task, presents the source selection criteria and construction pipeline, and summarizes the resulting data distribution.

\subsection{Task definition}
\label{sec:task}

Following prior work on semantic equivalence and contextual faithfulness~\citep{proofnetverif,faithalign}, we use \emph{alignment} to mean preservation of an informal target's mathematical content in its formalization.
This requires a mathematically justified correspondence of objects and representations that preserves the target's applicability conditions, substantive assumptions, and conclusions, or the concept specified by a definitional target.
LeanReviewBench evaluates this correspondence for selected declarations in Lean projects, using the source context and formal dependencies that determine their meaning.

\paragraph{Review input.}
Each review instance is a triple
\begin{equation*}
    x_t = (I_t, F_t, \mathcal{C}),
\end{equation*}
comprising an informal target $I_t$, its selected formal realization $F_t$, and a formal dependency context $\mathcal{C}$.
The informal component $I_t$ contains a source-grounded mathematical claim or an explicitly specified specialization, together with the source definitions, notation, and standing assumptions needed to read it independently of the surrounding article.
Informal extraction determines which source concepts require expansion and which serve as background mathematical concepts, making the chosen target's scope explicit.
The formal component $F_t$ consists of one or more selected Lean declarations, called \emph{roots}, that jointly express the target and receive a single alignment judgment.
It retains their elaborated types and, for definitions of mathematical objects or predicates, their defining values.
The context $\mathcal C$ supplies the dependencies needed to interpret these roots: project-defined predicates, structures, and quantitative expressions can determine mathematical content that is not explicit in the root types alone.

To construct $\mathcal C$, we follow declaration dependencies from the selected roots rather than include the entire repository, drawing on dependency-based context selection for formalization and semantic review~\citep{aria,leanatlas,shen2026mechalign}.
Starting from each root's elaborated type and, where applicable, its defining value, we collect the referenced declarations.
For each newly collected project declaration, we inspect its type and mathematical definition value in the same way, repeating until no further project dependencies are found.
This traversal includes predicate definitions and constructor types that expose inductive, structure, and class constraints.
Following the distinction between statement-relevant dependencies and theorem-proof dependencies used by Lean Compass~\citep{leanatlas}, we retain mathematical definitions while excluding proof bodies.
Proof bodies are replaced by \texttt{sorry} in the released Lean contexts; external-library references are retained with their identities and types, but are not expanded recursively.
This dependency-based context exposes mathematical content hidden behind the roots while keeping the review input focused, avoiding the incomplete information of root-only statements and the volume of unrelated material in a full repository.

\paragraph{Review output.}
The reviewer returns a binary prediction
\begin{equation*}
    \hat{y}_t \in \{\texttt{aligned},\ \texttt{misaligned}\}.
\end{equation*}
An \texttt{aligned} prediction states that $F_t$, interpreted through $\mathcal C$, preserves the mathematical content of $I_t$, allowing equivalent reformulations and representations.
A \texttt{misaligned} prediction identifies a substantive semantic discrepancy, including a strict strengthening or weakening of the target.
When giving the \texttt{misaligned} label, the reviewer must explain how the affected informal content differs from the relevant formal declarations.
Abstention is not an allowed benchmark output.

\subsection{Data Sources and Selection}
\label{sec:sources}

LeanReviewBench studies semantic alignment in contemporary mathematical formalization projects.
We select version-controlled Lean repositories with independent mathematical sources and project-local definitions, so that review can connect mathematical claims to their formal realizations and trace changes through development history.

We consider publicly accessible Lean repositories whose source revisions and licensing information can be recorded, using three selection criteria:
\begin{itemize}[itemsep=2pt, topsep=0pt, parsep=0pt, leftmargin=*]
\item \emph{Recent mathematical development.} The repository has substantive mathematical formalization development or review activity in 2026, beyond routine dependency, formatting, or CI maintenance. Earlier snapshots are recovered when tracing relevant changes.
\item \emph{Traceable development history.} Commits, issues, pull requests, or historical snapshots allow mathematical changes to be located and the affected declarations and their project context to be recovered.
\item \emph{Source-grounded project-level formalization.} The repository develops a coherent mathematical theory, result, or specification through related Lean declarations and project-local definitions, with an identifiable mathematical paper, blueprint, or independent specification as its source.
\end{itemize}

\begin{table}[t]
\centering
\footnotesize
\caption{Source repositories and mathematical categories. Roots counts distinct successfully extracted declaration names across versions; inputs counts successfully extracted versioned review tasks. Neither count represents independent mathematical cases. We collect 197 inputs from 143 roots in total.}
\label{tab:source-repo}
\begin{tabular*}{\linewidth}{@{\hspace{10pt}\extracolsep{\fill}} l l r r @{\hspace{10pt}}}
\toprule
Repository & Category & Roots & Inputs \\
\midrule
\nolinkurl{AlexKontorovich/PrimeNumberTheoremAnd} & Analytic number theory & 21 & 40\\
\nolinkurl{nikhgarg/AppliedModelingLib} & Applied modeling and economics & 26 & 30\\
\nolinkurl{teorth/pfr} & Additive combinatorics & 14 & 24\\
\nolinkurl{qinz1yang/differential-geometry} & Differential geometry and topology & 10 & 15\\
\nolinkurl{Verified-zkEVM/ArkLib} & Cryptographic protocols & 7 & 12\\
\nolinkurl{frenzymath/Poincare-Conjecture} & Riemannian geometry & 12 & 11\\
\nolinkurl{thefundamentaltheor3m/Sphere-Packing-Lean} & Discrete geometry & 6 & 9\\
\nolinkurl{fpvandoorn/carleson} & Harmonic analysis & 5 & 7\\
\nolinkurl{scottnarmstrong/CoarseDeGiorgi} & Regularity theory & 7 & 7\\
\nolinkurl{teorth/equational_theories} & Equational theories & 8 & 7\\
\nolinkurl{Robby955/FormalSLT} & Statistical learning & 4 & 6\\
\nolinkurl{leanprover-community/sphere-eversion} & Differential topology & 4 & 6\\
\nolinkurl{YaelDillies/apap} & Additive combinatorics & 2 & 6\\
\nolinkurl{optpku/stacks} & Stacks Project material & 6 & 5\\
\nolinkurl{ImperialCollegeLondon/FLT} & Number theory & 5 & 5\\
\nolinkurl{cameronfreer/vaught-conjecture-counterexample} & Model theory & 3 & 3\\
\nolinkurl{Lean-MoDS/StatsMLlib} & Statistics and learning & 2 & 2\\
\nolinkurl{RemyDegenne/brownian-motion} & Probability & 1 & 2\\
\midrule
\nolinkurl{Ivan-Sergeyev/seymour} & \multicolumn{3}{l}{Excluded: no usable inputs.} \\
\nolinkurl{anthropics/formal-math} & \multicolumn{3}{l}{Excluded: no usable inputs.} \\
\nolinkurl{emilyriehl/infinity-cosmos} & \multicolumn{3}{l}{Excluded: no usable inputs.} \\
\nolinkurl{girving/ray} & \multicolumn{3}{l}{Excluded: no usable inputs.} \\
\nolinkurl{leanprover-community/con-nf} & \multicolumn{3}{l}{Excluded: no usable inputs.} \\
\nolinkurl{YuanheZ/lean-stat-learning-theory} & \multicolumn{3}{l}{Excluded: migration to Lean 4.34.1 failed.} \\
\bottomrule
\end{tabular*}
\end{table}

Table~\ref{tab:source-repo} lists the selected repositories, their mathematical settings, and the results of input construction described in Section~\ref{sec:construction}.
We exclude several well-known repositories maintained by OpenAI and Anthropic because they lack a traceable development history.
The retained collection contains 197 versioned inputs and 143 distinct roots from 18 repositories; the remaining repositories are listed with their exclusion reasons.
Roots counts distinct declaration names within each repository across snapshots, while Inputs counts versioned review instances.
Historical versions of the same named declaration contribute multiple inputs, and a jointly reviewed input may contain multiple roots.

The sources span analysis, number theory, geometry, topology, probability, learning theory, applied modeling, cryptographic protocols, and equational theories.
Notable examples include projects on the Poincar\'e conjecture, characterizing the three-dimensional sphere~\citep{claypoincare}; Fermat's Last Theorem, ruling out positive-integer solutions to $a^n+b^n=c^n$ for integer $n>2$~\citep{fltproject}; eight-dimensional sphere packing, determining the densest arrangement of equal spheres~\citep{viazovska2017sphere}; and the prime number theorem, describing the asymptotic distribution of primes~\citep{nistpnt}.
Our inputs assess selected declarations within these projects.
Contributions range from two to 40 inputs per repository, with PrimeNumberTheoremAnd, AppliedModelingLib, and PFR together contributing 94 of the 197 inputs.
Across these settings, interpreting a selected result can require following analytic predicates, geometric structure constraints, quantitative bounds, or protocol specifications through project-local definitions.
The six excluded repositories yielded no usable inputs in this collection round or could not be migrated successfully to the common Lean environment, as recorded in Table~\ref{tab:source-repo}.

\subsection{Data construction}
\label{sec:construction}

Construction produces fixed review triples through historical discovery, independent target binding, reconstruction, dependency extraction, and reference review.
The following describes the general procedure; Figure~\ref{fig:construction-example} gives a simplified illustration using a FormalSLT adaptedness repair.
Implementation and validation details are provided in Appendix~\ref{app:implementation}.

\textbf{Historical discovery.}
We inspect author reports, repository documentation, commits, issues, and pull requests to identify candidate roots and recover their relevant source revisions.
When a corresponding revision is available, the snapshots are linked as a before/after pair; withdrawn declarations and other standalone candidates retain their individual snapshots.
These historical relationships guide discovery, while alignment is judged against the mathematical target bound to each input.
In the FormalSLT example, the commit title ``Fix vacuity in empirical-Bernstein CS'' leads us to inspect the increment-adaptedness change and recover the corresponding snapshots.

\textbf{Independent target binding.}
Each selected root is bound to an identifiable mathematical claim from a paper, blueprint, or independent specification.
The informal input retains the source definitions, notation, and standing assumptions required to interpret that claim.
When the target is a specialization of a source result, its chosen conditions and source correspondence are recorded explicitly.
The benchmark may use closely related source-grounded targets with different mathematical scopes; each selected target must remain faithful to its source or explicitly stated specialization.
Historical snapshots may therefore be bound to a shared target or to different targets, with alignment reviewed independently for each resulting input.
Historical records guide candidate discovery, while the selected informal target and formal realization determine the reference judgment through the annotation procedure described below.
For the illustrated pair, both snapshots are bound to a bounded-increment, predictable-variance specialization of Howard et al.'s Lemma~2 with a uniform mixing distribution; the target requires $X_k$ to be $\mathcal F_{k+1}$-measurable and $\mathbb E[X_k\mid\mathcal F_k]=0$.

\textbf{Historical reconstruction.}
We reconstruct the required historical sources under Lean~4.34.1 with pinned library dependencies.
Starting with the root's source module, we select its transitive project-local imports and migrate the selected modules to the common environment.
Compatibility edits preserve the historical statements and mathematical definitions and are recorded separately from upstream mathematical changes.
Candidates that cannot be reconstructed with their mathematical content intact are excluded.
In the illustrated pair, file-level selection reduces each snapshot from 119 Lean files to nine files containing about 80 source declarations, which are then migrated from Lean~4.30.0-rc2 to Lean~4.34.1.

\textbf{Dependency context extraction.}
LeanTrace~\citep{shen2026mechalign} extracts the selected roots and their declaration dependencies from the compiled environment according to Section~\ref{sec:task}.
File-level selection determines the reconstruction scope, while declaration-level extraction determines the formal content retained for review.
The packaged contexts are compiled and freshly extracted, and root types and necessary non-proof definition values are checked against the migrated, pre-packaging elaborated contexts.
In the example, declaration-level extraction further reduces the roughly 80 source declarations to about 10 project-local nodes in each dependency graph, with proof bodies omitted.
The retained adaptedness definitions expose the shift from $\mathcal F_k$-measurability to $\mathcal F_{k+1}$-measurability.

\textbf{Reference review and task packaging.}
GPT-6 Astra in Codex first examines the target binding, formal interpretation, and alignment under the criterion in Section~\ref{sec:task}, producing a proposed label with mathematical reasoning.
Inputs for which Astra cannot determine alignment with confidence are referred to a human annotator for a reference judgment.
A misalignment judgment includes a mathematical explanation tied to the informal target and relevant formal declarations.
The release contains binary reference annotations for all 197 inputs; diagnostic category adjudication and preparation of a frozen evaluation export are tracked separately.
Each review package combines $I_t$, $F_t$, and $\mathcal C$, with reference labels, review explanations, and discovery records stored separately.
Metadata link related versions and roots sharing a mathematical case; artifact details are given in Appendix~\ref{app:data-package}.
For the earlier FormalSLT snapshot, integrability and $\mathcal F_k$-measurability make the zero conditional-mean assumption force $X_k=0$ almost surely, providing the mathematical evidence for a restriction to degenerate increments.

\input{app/figure_eg_construct}

\subsection{Data distribution}
\label{sec:distribution}

The collection contains 197 versioned inputs, with binary reference labels of 87 \texttt{aligned} and 110 \texttt{misaligned}.
Figure~\ref{fig:data-distr} summarizes formal context size, misalignment types, and version pairing.

\textbf{Formal context size.}
We measure size by the number of project-local declaration nodes in each extracted graph, including roots and internal or generated declarations.
The 195 inputs shown in Figure~\ref{fig:data-distr-input-size} have a median of 10 nodes and a range of 1--425.
Two larger inputs, with 4,727 and 4,809 nodes, are omitted from the histogram.

\textbf{Misalignment types.}
We classify discrepancies against each input's bound informal target into five categories:
\emph{object and concept mismatch} (O), concerning the identity or interpretation of mathematical objects;
\emph{applicability mismatch} (A), concerning input domains or substantive premises;
\emph{quantification mismatch} (Q), concerning quantifier structure or witness dependence;
\emph{formula and quantitative mismatch} (F), concerning mathematical expressions or quantitative relations;
and \emph{property and coverage mismatch} (P), concerning required properties or conclusion coverage.
The 110 misaligned inputs have 134 category assignments: 87 have one category, 22 have two, and one has three.
Category counts overlap because an input may contain multiple discrepancies.
These diagnostic annotations remain provisional pending human category adjudication; annotation rules are detailed in Appendix~\ref{app:annotation}.

Applicability is the most frequent category, appearing in 48 inputs, followed by formula and quantitative mismatch in 30.
Property and coverage, object and concept, and quantification mismatches appear in 26, 22, and eight inputs, respectively.
The prominence of applicability and quantitative discrepancies highlights two aspects of project-level review: recovering conditions distributed across types, predicates, and structure fields, and checking the expressions that specify mathematical bounds and relations.

\textbf{Version pairing.}
The 69 historical pairs in Figure~\ref{fig:data-distr-pairing} cover 138 inputs (70.1\%); the remaining 59 (29.9\%) have no registered pair.
The pairs record actual development changes spanning root statements, definitions, and dependencies.
Each version is bound to its own informal target, so a pair may capture either a revision of the formalization against a shared target or a change in the intended mathematical scope.
Beyond independent alignment judgments, these relationships support future work on version comparison, semantic change localization, and repair review.

\pgfplotsset{
    layers/data distribution/.define layer set={
        axis background,axis grid,pre main,main,axis ticks,axis lines,
        axis tick labels,axis descriptions,axis foreground
    }{/pgfplots/layers/standard}
}
\begin{figure}[t]
    \centering
    \begin{subfigure}[t]{0.36\linewidth}
        \centering
        \begin{tikzpicture}
            \begin{axis}[
                set layers=data distribution,
                width=\linewidth,
                height=4cm,
                scale only axis=false,
                ybar interval,
                xmode=log, log basis x=2,
                xmin=1, xmax=512,
                ymin=0, ymax=40,
                x tick label as interval=false,
                xtick={2,4,8,16,32,64,128,256,512},
                xticklabels={2,4,8,16,32,64,128,256,512},
                xticklabel style={rotate=0},
                ytick={0,10,20,30,40},
                xlabel={Project declarations per input},
                ylabel={Number of inputs},
                tick label style={font=\scriptsize},
                label style={font=\scriptsize},
                axis lines=left,
                axis line style={black!65},
                tick align=outside,
                ymajorgrids=true,
                grid style={black!10},
                clip=false,
            ]
                \addplot[fill=blue!45, draw=white, line width=0.3pt]
                coordinates {
                    (1, 28)
                    (2, 18)
                    (4, 31)
                    (8, 35)
                    (16, 24)
                    (32, 30)
                    (64, 18)
                    (128, 2)
                    (256, 9)
                    (512, 0)
                };
            \end{axis}
        \end{tikzpicture}
        \subcaption{Input size (base-2 logarithmic bins).}
        \label{fig:data-distr-input-size}
    \end{subfigure}
    \hfill
    \begin{subfigure}[t]{0.34\linewidth}
        \centering
        \begin{tikzpicture}
            \begin{axis}[
                set layers=data distribution,
                width=\linewidth, height=4.2cm,
                ybar, bar width=16pt,
                enlarge x limits=false,
                symbolic x coords={O,A,Q,F,P},
                xmin={[normalized]-0.5}, xmax={[normalized]4.5},
                xtick=data, ymin=0, ymax=55,
                ytick={0,10,20,30,40,50},
                xlabel={Misalignment category},
                ylabel={Number of inputs},
                tick label style={font=\scriptsize},
                label style={font=\scriptsize},
                nodes near coords,
                every node near coord/.append style={font=\scriptsize},
                axis lines=left, axis line style={black!65},
                tick align=outside, ymajorgrids=true,
                grid style={black!10},
            ]
                \addplot[fill=blue!45, draw=white]
                    coordinates {(O,22) (A,48) (Q,8) (F,30) (P,26)};
            \end{axis}
        \end{tikzpicture}
        \subcaption{Misalignment types (multi-label).}
        \label{fig:data-distr-categories}
    \end{subfigure}
    \hfill
    \begin{subfigure}[t]{0.26\linewidth}
        \centering
        \begin{tikzpicture}[font=\scriptsize]
            \path[use as bounding box] (-1.5,-2) rectangle (1.5,2);
            \fill[blue!45] (0,0) -- (0:1)
                arc[start angle=0,end angle=252.182741,radius=1] -- cycle;
            \fill[orange!55] (0,0) -- (252.182741:1)
                arc[start angle=252.182741,end angle=360,radius=1] -- cycle;
            \draw[white, line width=0.8pt] (0:0.63) -- (0:1);
            \draw[white, line width=0.8pt] (252.182741:0.63) -- (252.182741:1);
            \fill[white] (0,0) circle[radius=0.63];
            \node[align=center] at (0,0) {\textbf{197}\\inputs};
            \node[align=center] at (0,1.48)
                {\textcolor{blue!65!black}{Paired}\\138 (70.1\%)};
            \node[align=center] at (0,-1.48)
                {\textcolor{orange!65!black}{Unpaired}\\59 (29.9\%)};
        \end{tikzpicture}
        \subcaption{Version pairing.}
        \label{fig:data-distr-pairing}
    \end{subfigure}
    \caption{Benchmark data distribution. Panel (a) counts project-local
    declaration nodes, including roots, and excludes external-library boundary
    nodes. The two Equational Theories outliers, \texttt{etp-endtoend} and
    \texttt{etp-law-coverage}, are omitted from this panel only. Panel (b) counts inputs containing each category;
    multiple categories may apply to one input. O: object and concept;
    A: applicability; Q: quantification; F: formula and quantitative;
    P: property and coverage. Panel (c) counts inputs by registered pairing status;
    the 138 paired inputs form 69 before/after pairs.}
    \label{fig:data-distr}
\end{figure}
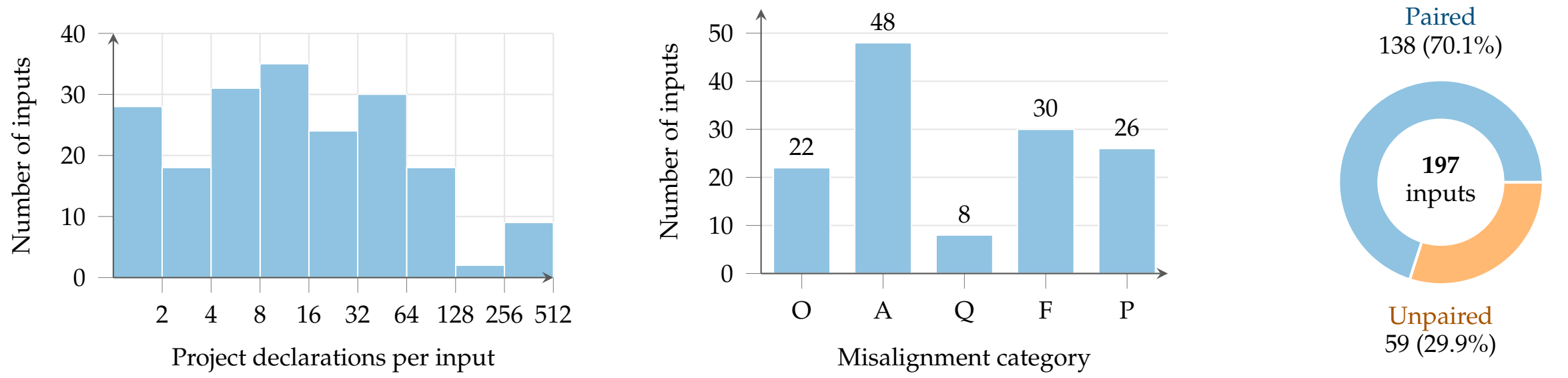


\section{Evaluation}
\label{sec:evaluation}
We evaluate six models under two prompting strategies to assess their ability to review semantic alignment in project-level Lean formalizations. The evaluation measures accuracy across all inputs and classification performance on valid binary judgments against the human-reviewed reference labels. We first describe the experimental setup, then present the results and analyze the reliability of model judgments.

\subsection{Experimental Setup}

We evaluate all 197 inputs defined in Section~\ref{sec:task} against their final human-reviewed binary labels.
Each input supplies the informal target with its scope qualifications and supporting mathematical context, the selected Lean declarations, and their formal dependencies.
The formal context is presented as structured text without explicit dependency-graph edges.
Theorem proof bodies and reference labels are excluded from the model input.

\textbf{LLMs.}
We evaluate six models in three groups: large models (GPT-6.1 Sol~\citep{gpt61sol}, Qwen3.8 Max~\citep{qwen38max}, and Kimi K3.1 Pro~\citep{kimik31pro}), small models (Qwen3.8 27B~\citep{qwen3827b} and Gemma 4 E4B~\citep{gemma4}), and a decision model (Jev 1.14~\citep{jev114}).
These groups allow us to compare general-purpose models at different scales and examine a model specialized for judgment tasks.
We follow each model's original configuration, including its context-length limit, without manually truncating the inputs.
Model versions and inference settings are provided in Appendix~\ref{app:evaluation-details}.

\textbf{Prompts.}
Each input is evaluated independently under two prompting strategies.
Direct provides a minimal instruction to judge faithfulness without prescribing a comparison procedure.
ReForm~\citep{reform} uses the original released prompt without modification and requests a structured analysis of the informal and formal statements.
Outputs from both strategies are converted to the same binary alignment labels.
Complete prompt templates and output parsing rules are provided in Appendix~\ref{app:evaluation-details}.

\textbf{Metrics.}
We report accuracy on all 197 inputs as the fraction whose predicted labels match the human-reviewed references, counting failures as incorrect. 
Inputs for which an \texttt{aligned} or \texttt{misaligned} label is successfully obtained are treated as valid samples; their counts are reported in the table. 
On these valid samples, we further compute precision, recall, and F1, treating \texttt{misaligned} as the positive class. 
Precision is the fraction of inputs predicted as misaligned that are misaligned according to the reference labels; recall is the fraction of reference-misaligned inputs that are predicted as misaligned. 
F1 is the harmonic mean of precision and recall. Details are provided in Appendix~\ref{metric_details}. 
Failures are excluded from the confusion matrix and from these three metrics, so different configurations may be evaluated on different subsets of inputs.

\subsection{Results}
\label{sec:results}
Table~\ref{baseline} reports results for six models under Direct and ReForm prompting on all 197 inputs against the final human-reviewed labels.
It presents accuracy and valid-output counts, followed by precision, recall, and F1 computed on valid binary judgments.

\begin{table}[t]
\centering
\small
\setlength{\tabcolsep}{4pt}
\renewcommand{\arraystretch}{1.15}

\caption{
Comparison of prompting strategies on 197 evaluation instances
using final human-reviewed reference labels.
Accuracy uses all 197 inputs. Valid output counts inputs with a valid binary
judgment. Precision, recall, and F1 are computed on those inputs, with
\texttt{misaligned} as the positive class. Bold indicates the higher value
for each model and classification metric.
}
\label{tab:prompt-comparison}
\label{baseline}

\begin{tabular*}{\linewidth}{@{\hspace{10pt}\extracolsep{\fill}} l l c c c c c @{\hspace{10pt}}}
\toprule
\multirow{2}{*}{Model}
& \multirow{2}{*}{Prompt}
& \multicolumn{2}{c}{All inputs (197)}
& \multicolumn{3}{c}{Valid binary judgments} \\
\cmidrule(lr){3-4}\cmidrule(lr){5-7}
& & Accuracy & Valid output & Precision & Recall & F1 \\
\midrule

\multirow{2}{*}{GPT-6.1 Sol}
& Direct & \textbf{0.827} & 196 & \textbf{0.881} & 0.809 & \textbf{0.844} \\
& ReForm & 0.746 & 194 & 0.721 & \textbf{0.927} & 0.811 \\

\cmidrule(lr){1-7}
\multirow{2}{*}{Qwen3.8 Max}
& Direct & 0.614 & 193 & \textbf{0.821} & 0.426 & 0.561 \\
& ReForm & \textbf{0.650} & 191 & 0.800 & \textbf{0.556} & \textbf{0.656} \\

\cmidrule(lr){1-7}
\multirow{2}{*}{Kimi K3.1 Pro}
& Direct & 0.701 & 196 & \textbf{0.871} & 0.555 & 0.678 \\
& ReForm & \textbf{0.777} & 192 & 0.863 & \textbf{0.759} & \textbf{0.808} \\

\cmidrule(lr){1-7}
\multirow{2}{*}{Qwen3.8 27B}
& Direct & \textbf{0.609} & 166 & \textbf{0.795} & 0.652 & 0.716 \\
& ReForm & 0.533 & 141 & 0.742 & \textbf{0.697} & \textbf{0.719} \\

\cmidrule(lr){1-7}
\multirow{2}{*}{Gemma 4 E4B}
& Direct & \textbf{0.492} & 188 & 0.710 & 0.212 & 0.326 \\
& ReForm & 0.137 & 42 & \textbf{0.789} & \textbf{0.577} & \textbf{0.667} \\

\cmidrule(lr){1-7}
\multirow{2}{*}{Jev 1.14}
& Direct & 0.645 & 195 & \textbf{0.828} & 0.482 & 0.609 \\
& ReForm & \textbf{0.650} & 190 & 0.667 & \textbf{0.819} & \textbf{0.735} \\

\bottomrule
\end{tabular*}
\end{table}
Direct prompting achieves the highest reported accuracy (0.827), followed by Kimi K3.1 Pro with ReForm (0.777).
On valid binary judgments, GPT-6.1 Sol with Direct prompting achieves the highest reported F1 (0.844) and precision (0.881), while its ReForm configuration achieves the highest recall (0.927). Valid-output counts range from 141 to 196 for all configurations except Gemma 4 E4B with ReForm, which returns 42 valid labels.
ReForm produces fewer valid outputs than Direct for six models.
Classification quality and the ability to return a valid judgment both affect overall accuracy: Gemma 4 E4B's low accuracy under ReForm reflects both missing judgments and classification errors.
The aggregate results do not establish whether instruction following, context limits, or other factors caused the missing outputs.
Precision, recall, and F1 use each configuration's valid outputs, which may cover different subsets of inputs.

\subsection{Analysis}
\label{sec:analysis}
\subsubsection{Prompt Sensitivity}
For each model, Direct and ReForm use the same mathematical inputs and inference configuration but differ in their reasoning requirements and output formats. Their effects on performance vary across models. ReForm achieves higher accuracy for Kimi K3.1 Pro, Qwen3.8 Max, and Jev 1.14, whereas Direct performs better for GPT-6.1 Sol, Qwen3.8 27B, and Gemma 4 E4B. In terms of F1, ReForm outperforms Direct for five of the six models, with GPT-6.1 Sol being the only exception. The improvement is particularly pronounced for Gemma 4 E4B, while the difference for Qwen3.8 27B is marginal. These results suggest that the more structured ReForm prompt can improve misalignment detection for several models, although its effect on overall accuracy depends on the model.

The relationship between accuracy and classification performance also varies across models. For Qwen3.8 27B and Gemma 4 E4B, Direct achieves higher overall accuracy, while ReForm obtains higher F1 on valid binary judgments. This difference reflects the distinct evaluation scopes: accuracy includes all inputs and counts unsuccessful outputs as incorrect, whereas precision, recall, and F1 are computed only on inputs with valid binary judgments. The distinction is particularly important for Gemma 4 E4B, whose valid-output count drops from 188 under Direct to 42 under ReForm. Although its F1 increases from 0.326 to 0.667, this result is based on a substantially smaller subset of inputs and should not be interpreted as an improvement in performance across the full evaluation set.

Prompt choice also affects the balance between detecting misalignment and avoiding false alarms. For GPT-6.1 Sol, Qwen3.8 Max, and Jev 1.14, ReForm increases recall while reducing precision. In a screening workflow where predictions of misalignment trigger further review, false negatives may allow semantic discrepancies to pass undetected, whereas false positives primarily incur additional review costs. When the cost of missed discrepancies is substantially higher, the increased recall under ReForm may be desirable despite a reduction in precision. Overall, these results indicate that prompt selection involves a trade-off between classification performance, output reliability, and the relative costs of false positives and false negatives.

\subsubsection{Reliability of Judgment Reasons.}
Agreement with a binary reference label does not, by itself, establish that a model has correctly identified the semantic relationship between the informal target and the formalization.
We therefore examine whether correct judgments are supported by valid and sufficient reasons.
Reliable reasons can support subsequent confirmation, localization, and repair of semantic discrepancies.
Among the 109 available ReForm responses from GPT-6.1 Sol on cases labeled misaligned in LeanReviewBench, 101 agreed with the reference labels. A GPT-6 Astra review found that 92 of these responses gave reasons consistent with the reference rationales. These proportions describe the reviewed correct predictions for this configuration, rather than explanation reliability across all responses or models.

On \texttt{arklib:sumcheck-full-rbr-after}, GPT-6.1 Sol under ReForm predicts \texttt{misaligned}, agreeing with the reference label.
However, its explanation explicitly dismisses the algebraic mismatch, claiming that ``a finite commutative domain semiring admits field structure.''
This claim is false for the formal assumptions \texttt{CommSemiring R}, \texttt{Fintype R}, and \texttt{IsDomain R}: the last supplies nontriviality and cancellation by nonzero multiplicative factors, but does not supply additive inverses.

Consider the Boolean semiring
\[
  B=\{0,1\}, \qquad
  a+b := a\lor b, \qquad
  a\cdot b := a\land b.
\]
It is a finite commutative semiring satisfying the stated cancellation condition, since its only nonzero element is the multiplicative identity.
Nevertheless, $1+x=1$ for every $x\in B$, so $1$ has no additive inverse.
Thus, these operations cannot extend to a field structure.

The source requires a finite integral domain, whereas the formal assumptions admit this additional structure.
Under the benchmark's applicability criterion, this domain difference supports the reference verdict. The model instead bases its rejection on concerns about the security predicate while incorrectly ruling out the reference diagnosis.
This example demonstrates that a correct binary verdict can coexist with an explicit mathematical error in the explanation.
The Boolean semiring separates the admissible domains; it is not a counterexample to the protocol's soundness bound.

\section{Conclusion}

LeanReviewBench evaluates whether selected Lean declarations, interpreted through their project dependencies, preserve the mathematical content of independently sourced informal targets.
It contains 197 versioned inputs from 18 contemporary projects, with 87 aligned and 110 misaligned reference labels and 69 historical pairs.
Our evaluation of six models under Direct and ReForm prompting provides baselines for semantic review of existing formalizations, measuring both accuracy over all inputs and classification performance on valid binary outputs.
The reported results show variation across models and prompting strategies, with output reliability also affecting overall performance.

The evaluation has several limitations.
Without a context ablation, it does not establish the contribution of the supplied dependencies.
Related historical versions and uneven project contributions also limit how aggregate scores generalize across projects.
Future work should compare context settings and report results by project and mathematical case group.
Reviewing the validity and sufficiency of model-provided reasons alongside label correctness would further assess whether automated judgments provide reliable mathematical evidence.

\section*{Acknowledgment}

This paper is supported by the National Natural Science Foundation of China (Grant No. 12288201) and the Strategic Priority Research Program of CAS (Grants XDA0480501 and XDA0480502).

\bibliographystyle{KLMM/klmm}
\bibliography{refs}

\clearpage
\appendix
\section{Construction, Annotation, and Release Details}
\label{app:implementation}

This appendix supplements the construction procedure in Section~\ref{sec:construction} with migration and validation details, annotation rules, and the release layout.

\subsection{Construction and Validation}
\label{app:construction-validation}

Historical snapshots are reconstructed under the common environment in Table~\ref{tab:app-environment}; per-input upstream revisions and mathematical-source locators are recorded separately.
Migration is limited to compatibility adaptations that preserve historical statements and mathematical definitions, such as interface and import updates, explicit elaboration arguments, and limited interface support.
Candidates requiring substantive mathematical redevelopment are abandoned.
Some contexts retain historical interfaces, including Carleson's norm and measurability definitions, ArkLib's Cslib compatibility modules, and FLT's completion wrappers.
Per-input migration notes document these adaptations separately from upstream mathematical changes; the restriction concerns their nature rather than a fixed number of affected files.

Dependency extraction follows the traversal described in Section~\ref{sec:task}.
The packaged slices additionally retain notation, aliases, instances, and runtime resources needed for elaboration.
Proof-valued bodies are replaced by \texttt{sorry}; computational definitions and data-producing tactics are preserved.
All 197 contexts have recorded successful compilation and fresh dependency extraction.
Root types and necessary non-proof definition values are compared with the migrated, pre-packaging environment, allowing proof irrelevance and source-neutral binder and let annotations.
These comparisons validate slicing and proof erasure, rather than equivalence to the original dependency environment.
Original-toolchain elaboration was not checked for the 26 added inputs, and the checks do not certify erased proofs or semantic alignment.

\begin{table}[h]
\centering
\small
\caption{Common reconstruction environment (Lean 4.34.1). Per-input upstream revisions are recorded separately.}
\label{tab:app-environment}
\begin{tabular*}{\linewidth}{@{\hspace{5pt}\extracolsep{\fill}} l l l @{\hspace{5pt}}}
\toprule
Component & GitHub Repo & GitHub commit hash \\
\midrule
Lean & \texttt{leanprover/lean4} & \nolinkurl{5045d0056413266e57c625dcd7c365b10e377c52} \\
Mathlib & \texttt{leanprover-community/mathlib4} & \nolinkurl{d13f23b723b8a846827a245b89c10fc7d3f11612} \\
Cslib & \texttt{leanprover/cslib} & \nolinkurl{7c8f6c0f67015df29b152fdcffaf7850db2e9185} \\
\bottomrule
\end{tabular*}
\end{table}

\subsection{Reference Annotation}
\label{app:annotation}

We apply the alignment criterion in Section~\ref{sec:task} using GPT-6 Astra in Codex for initial annotation. For inputs associated with identifiable repair commits, we use historical changes as review signals: pre-repair versions are expected to be \texttt{misaligned}, while post-repair versions are expected to be \texttt{aligned}. GPT-6 Astra in Codex provides initial judgments. Cases where Astra's predictions contradict these expected directions, as well as cases classified as \texttt{undetermined}, are referred to a human annotator for further review. Human annotators examine the mathematical sources and formal declarations to determine the reference judgments for all cases requiring human review, including cases without identifiable historical repair information. Negative judgments identify the affected informal content and relevant formal declarations; historical repair reports and before/after positions do not determine labels.

\begin{table}[ht]
\centering
\small
\caption{Diagnostic categories and their principal boundaries. Categories apply to independent discrepancies against the bound informal target.}
\label{tab:app-taxonomy}
\begin{tabular*}{\linewidth}{@{\hspace{5pt}\extracolsep{\fill}} l l m{320pt} @{\hspace{5pt}}}
\toprule
ID & Category & Inclusion and boundary rule \\
\midrule
O & Object and concept & A different mathematical object or concept, or a missing binding to the designated object; distinguish from a local expression error. \\
\cmidrule(lr){1-3}
A & Applicability & A changed input domain or substantive premise, including restrictions encoded in types, instances, or structure fields. \\
\cmidrule(lr){1-3}
Q & Quantification & A changed quantifier kind, order, scope, or witness dependence; general domain restrictions belong to A. \\
\cmidrule(lr){1-3}
F & Formula and quantitative & A changed operation, coefficient, sign, exponent, index, normalization, or quantitative relation within the intended object. \\
\cmidrule(lr){1-3}
P & Property and coverage & A missing, added, or replaced output property or independent conclusion; distinguish output guarantees from input premises. \\
\bottomrule
\end{tabular*}
\end{table}

The diagnostic categories in Table~\ref{tab:app-taxonomy} describe independent discrepancies against the bound target, including its qualifications.
Each input receives at most three distinct categories; multiple discrepancies in one category count once, and consequences such as vacuity do not receive an additional category solely for that consequence.
Counts use versioned inputs, so related versions may repeat a discrepancy.
The standard and input-bound annotations are retained in \texttt{metadata/taxonomy/}.

\subsection{Released Artifacts and Reproducibility}
\label{app:data-package}

The release contains 197 inputs in 18 source batches using the native \texttt{paperlean-project-case-1} format.
Listing~\ref{lst:app-artifacts} describes the input packages and accompanying records.
Input and formal-source hashes bind validation results and reference evidence to the corresponding artifacts; offline validation checks these bindings and coverage without recompiling Lean.
The released Lean files are authoritative, and omitted dependency graphs can be regenerated under the pinned environment.

\begin{lstlisting}[
  language={},
  basicstyle=\ttfamily\footnotesize,
  columns=fullflexible,
  keepspaces=true,
  showstringspaces=false,
  breaklines=true,
  frame=single,
  rulecolor=\color{black!25},
  xleftmargin=4pt,
  xrightmargin=4pt,
  caption={Release directory structure and record contents. Curator metadata require filtering before model use.},
  label={lst:app-artifacts},
  captionpos=b
]
datasets/
|-- project-cases.json       // Source batches and pinned environment
`-- <batch>/
    |-- README.md            // Source description and scope
    |-- index.csv            // Versioned-input inventory
    `-- projects/<id>/
        |-- project.json     // Revision, roots, modules, version/group links,
        |                    // technical states, and input hashes
        |-- informal.jsonl   // Targets, sources, locators, qualifications,
        |                    // and bindings to formal roots
        |-- CHANGES.md       // Migration and packaging changes
        `-- formal/
            |-- Main.lean    // Selected roots and local imports
            `-- dep/         // Proof-erased dependency slices and support
                             // Required runtime resources are also retained
metadata/
|-- labels/                  // Binary labels and concise reasons
|-- taxonomy/                // Standard and provisional diagnostic labels
|-- evidence/
|   |-- upstream/            // Upstream discovery claims
|   |-- analysis/            // Semantic review evidence and binding index
|   `-- technical/           // Compact checks for added source batches
|-- provenance.json          // Source and historical preparation locators
|-- materials/               // Independent-source locators and byte hashes
|-- validation.json          // Compilation, extraction, preservation checks;
|                            // input/formal digests and selected roots
|-- manifest.json            // Data, metadata, and license hash inventory
`-- support-inventory.json   // Documentation, configuration, script/test hashes
licenses/                    // Source attribution and license evidence
\end{lstlisting}

Native packages require a separate model-input export to exclude curator metadata and version clues.
The export retains the informal target and qualifications, selected roots, and declarations needed to interpret them, while removing proof bodies, Lean comments, reference labels and reasons, review reports, and discovery provenance.
Declaration identities needed for mathematical interpretation remain available.
The JSON presentation in Appendix~\ref{app:evaluation-details} supplies declaration contents without explicit graph edges.
The native release is not itself a frozen scoring export: current evaluation tools consume separately prepared evaluation packages.

\section{Evaluation Details}
\label{app:evaluation-details}

\subsection{Input presentation and run protocol.}
Each evaluation instance is serialized as a structured JSON
object and incorporated into the user message together with
the task instructions. Thus, the model receives a single
user message combining the evaluation prompt and the
JSON-formatted input, rather than separate system and
user messages. The JSON input consists of five top-level
fields:

\begin{itemize}[itemsep=2pt, topsep=0pt, parsep=0pt, leftmargin=*]
    \item task\_id: A unique identifier for the
    evaluation instance.

    \item main\_theorem: The target theorem,
    including its informal mathematical statement, formal
    Lean statement, and supplementary qualifications.

    \item definitions: Auxiliary definitions
    associated with the target, when applicable.

    \item formal\_context: Relevant formal
    declarations, including their identifiers, declaration
    kinds, types, and available definition values, together
    with metadata for external declarations.

    \item context: The Lean environment
    configuration, including the toolchain version,
    standalone configuration, and external dependency
    boundary.
\end{itemize}

The input exposes the contents of relevant declarations
but does not explicitly provide dependency-graph edges
or the proof bodies of the target theorem. Reference
labels are excluded from all model-visible inputs.
A representative model input is provided in
Appendix~\ref{app:user-message}.
\begin{table*}[tbp]
  \centering
  \caption{Input length in tokens for the same 197 examples under Direct
    and ReForm. We tokenize each message body with o200k\_base
    and sum the counts within each request.}
  \label{tab:direct-input-tokens}
  \small
  \begin{tabular*}{\linewidth}{@{\extracolsep{\fill}}lrrrrrr@{}}
    \toprule
    Configuration & $n$ & Mean & Median & Min & Max & P90 \\
    \midrule
    Direct & 197 & 40,582.21 &  7,289 &   537 & 3,319,343 &  66,602 \\
    ReForm & 197 & 41,073.51 & 7,782 & 1,029 & 3,319,836 & 67,093 \\
    \bottomrule
  \end{tabular*}
\end{table*}
We further measure the input length of each evaluation
instance under both Direct and ReForm prompting strategies.
We tokenize the complete user-message body using
o200k\_base and report the token counts across
the same 197 evaluation instances in
Table~\ref{tab:direct-input-tokens}.

\subsection{Model versions and inference settings.}
We evaluated GPT-6.1 Sol with medium reasoning effort, Qwen3.8 Max with thinking enabled and a budget of 8,192 thinking tokens, and Kimi K3.1 Pro with
high reasoning effort. The officially documented context windows are 1,050,000,
1,000,000, and up to 1,048,576 tokens, respectively. Qwen3.8-27B was
served through a self-hosted OpenAI-compatible API with
thinking enabled and medium reasoning effort. Four NVIDIA
A800 80~GB GPUs each hosted an independent replica, with
one GPU per request. Gemma 4 E4B ran on vLLM with one replica
on a single NVIDIA H200 NVL GPU, supporting up to two
concurrent requests; other sampling and thinking settings
remained at server defaults. Both deployments used BF16
precision and a 131,072-token context limit. Jev 1.14 was evaluated through TypeSafe's official System One
API using one binary Choice question per case and prompt,
returning the selected label, option probabilities, and
confidence. Default context limits were 64K tokens per
request and 32K per question, with the latter covering
shared state plus the question. Only explicit context-limit
rejections triggered an identical retry in long-context mode,
which raised both limits to 262K tokens. All models except Jev used a maximum completion budget of
12,288 tokens per request; for the two local deployments,
this included reasoning and final-answer tokens. Inputs to
the local models and Jev were never truncated, and context
overflows were recorded as failures after any applicable
long-context retry. All models used a standard per-case
timeout of 420~s.

\subsection{Output parsing and failure accounting.}
Direct requests an \texttt{aligned} or \texttt{misaligned} label. ReForm requests a JSON object whose \texttt{is\_assistant\_correct} field contains Correct or Incorrect; these judgments are mapped to \texttt{aligned} and \texttt{misaligned}, respectively. If no valid binary label can be obtained, the input is recorded as a failure. Failures are excluded from TP, FP, TN, and FN when computing precision, recall, and F1, but remain in the denominators of accuracy and failure rate. Possible causes include not following the output instructions or reaching the model context limit without successfully producing a judgment.

Direct responses are accepted only when the entire response, after trimming surrounding whitespace and converting to lowercase, is exactly aligned or misaligned. ReForm responses are parsed as JSON objects, optionally enclosed in a complete Markdown code fence, with a string-valued reasons field and an \texttt{is\_assistant\_correct} field equal to the case-sensitive string Correct or Incorrect. These values map to aligned and misaligned, respectively. Duplicate keys and malformed JSON are rejected; labels are neither inferred from explanations nor recovered by repairing malformed responses.

\subsection{Metric Details}
\label{metric_details}

The confusion matrix below summarizes the four
possible classification outcomes, treating
\texttt{misaligned} as the positive class.
All counts are defined relative to the
human-reviewed reference labels and computed
over valid evaluation samples.

\begin{center}
\small
\setlength{\tabcolsep}{10pt}
\renewcommand{\arraystretch}{1.2}
\begin{tabular}{@{}llcc@{}}
    \toprule
    & & \multicolumn{2}{c}{\textbf{Predicted label}} \\
    \cmidrule(lr){3-4}
    & & \texttt{misaligned} $(+)$
      & \texttt{aligned} $(-)$ \\
    \midrule
    \multirow{2}{*}{\textbf{Reference label}}
    & \texttt{misaligned} $(+)$ & TP & FN \\
    & \texttt{aligned} $(-)$ & FP & TN \\
    \bottomrule
\end{tabular}
\end{center}

True positives (TP) and true negatives (TN)
correspond to correctly classified misaligned
and aligned samples, respectively.
False positives (FP) are aligned samples
incorrectly predicted as misaligned, whereas
false negatives (FN) are misaligned samples
incorrectly predicted as aligned.

Precision ($P$), recall ($R$), and F1 ($F_1$)
are then defined as
\begin{align}
P &= \frac{\mathrm{TP}}{\mathrm{TP}+\mathrm{FP}},
\label{eq:precision}
\\[4pt]
R &= \frac{\mathrm{TP}}{\mathrm{TP}+\mathrm{FN}},
\label{eq:recall}
\\[4pt]
F_1 &= \frac{2PR}{P+R}
     = \frac{2\mathrm{TP}}
            {2\mathrm{TP}+\mathrm{FP}+\mathrm{FN}}.
\label{eq:f1}
\end{align}

Precision measures the correctness of predicted
misalignments, while recall measures the coverage
of reference misalignments.
The F1 score summarizes both through their
harmonic mean.

\subsection{System Prompts}
\label{app:prompts}

\begin{promptbox}{Direct Prompt}
Decide whether the supplied Lean formal target faithfully states the supplied informal mathematical claim. Return only aligned or misaligned.
\end{promptbox}

\begin{promptbox}{ReForm Prompt}
\textbf{Role:} Lean \& Formal Verification Expert

\medskip
\textbf{Input:}
\begin{itemize}
    \setlength{\itemsep}{2pt}
    \item Mathematical\_Text: A math problem and its answer
    (no proof).
    \item Lean4Code: A Lean 4 theorem statement formalizing
    the problem. Proof is intentionally omitted (e.g., \texttt{sorry}).
\end{itemize}

\textbf{Goal:}

Determine if the Lean theorem statement is an exact and faithful
formalization of the mathematical problem.

\textbf{Do not evaluate or consider the answer or the proof.
Your sole task is to verify the correctness of the formalization.}

\medskip
\textbf{Evaluation Stages (All required):}

\medskip
\textbf{1. Math Assertion Analysis}

Identify all structurally and semantically relevant components of the
mathematical problem, including variables, types, quantifiers,
constraints, logic structure, conclusion, and so on.
The analysis should be based on the actual content of the text.

\medskip
\textbf{2. Lean Statement Analysis (ignore proof part)}

Extract all structurally and semantically relevant components from the
Lean statement, including variables, types, conditions, quantifiers,
constraints, the final claim, and so on.
The analysis should reflect the actual content present in the Lean code.

\medskip
\textbf{3. Comparative Verification}

Check for exact correspondence between the math and Lean statements;
you may refer to aspects like:
\begin{itemize}
    \setlength{\itemsep}{2pt}
    \item Semantic alignment, logic structure, and quantifier correctness.
    \item Preservation of constraints and boundary assumptions.
    \item Accurate typing and use of variables.
    \item Syntactic validity and proper Lean usage (free from errors).
    \item Use of symbols and constructs without semantic drift.
    \item No missing elements, no unjustified additions, and no
    automatic corrections or completions.
\end{itemize}

\textbf{4. Final Judgement}

Based solely on the above analysis, judge whether the Lean statement
is a correct and exact formalization of the mathematical problem.

\medskip
\textbf{5. Accuracy Confirmation}

\textbf{If correct:} clearly confirm why all elements match.

\textbf{If incorrect:} list all mismatches and explain how each one
affects correctness.

\medskip
\textbf{Note:}
While the analysis may be broad and open to interpreting all relevant
features, the final judgment must be based only on what is explicitly
and formally expressed in the Lean statement.

\textbf{Do not consider or assess any part of the proof.
Your judgment should be entirely about the accuracy of the statement
formalization.}

\medskip
\textbf{Output Format:}

Return exactly one JSON object:
\begin{lstlisting}[style=promptjson]
{
  "reasons": "Your detailed CoT analysis:
1. Math Assertion Analysis: [...]
2. Lean Statement Analysis (Proof Ignored): [...]
3. Comparative Verification: [...]
4. Conclusion: [...]
5. Accuracy Confirmation: [...match confirmation or list of discrepancies...]",
  "is_assistant_correct": "[Correct/Incorrect]"
}
\end{lstlisting}
\end{promptbox}

\subsection{Json Templates}
\label{app:user-message}

We present an input json example as following.

\begin{promptbox}{Json Example}
\begin{lstlisting}[style=promptjson,
    literate={∃}{{\ensuremath{\exists}}}1
             {∧}{{\ensuremath{\wedge}}}1
             {¬}{{\ensuremath{\neg}}}1
             {→}{{\ensuremath{\rightarrow}}}1
             {∀}{{\ensuremath{\forall}}}1
             {◇}{{\ensuremath{\diamond}}}1
             {α}{{\ensuremath{\alpha}}}1]
{
  "task_id": "case-4f3d9e0a9bd151c17d14895a",
  "main_theorem": {
    "id": "case-4f3d9e0a9bd151c17d14895a", "kind": "theorem",
    "formal": {
      "type": "∃ G x, Equation1729 G ∧ ¬Equation817 G",
      "value": null, "proof_valued": true, "projection": false, "private": false, "generated_kind": null,
      "statement_source": "theorem Eq1729.not_817 : ∃ G x, Equation1729 G ∧ ¬Equation817 G"
    },
    "informal": "\\begin{theorem}[1729 does not imply 817]\\label{1729_refute_817}\\lean{Eq1729.not_817}\\leanok There exists a magma that satisfies equation 1729 but not equation 817.\n\\end{theorem}\n\nEquational laws (independent paper numbering table; universally quantified magma variables):\n    \\x &\\formaleq \\x \\op ((\\x \\op \\x) \\op (\\x \\op \\x)) \\label{eq817}\\tag{E817} \\\\\n    \\x &\\formaleq (\\y \\op \\y) \\op ((\\y \\op \\x) \\op \\y) \\label{eq1729}\\tag{E1729} \\\\",
    "qualifications": ["Full greedy chain and extension obligations are retained, including SmallMagma, Edit, MagmaConstruction and ExtensionTheorem. It is a named coherent mathematical construction rather than a singleton exercise."]
  },
  "definitions": [],
  "formal_context": {
    "declarations": [
      {"id": "Equation1729", "kind": "abbrev", "module": "dep.equational_theories.Equations.Eqns1000_1999",
       "formal": {"type": "(G : Type u) → [a : Magma G] → Prop",
         "value": "fun G [Magma G] => ∀ (x y : G), x = (y ◇ y) ◇ ((y ◇ x) ◇ y)",
         "proof_valued": false, "projection": false, "private": false, "generated_kind": null}},
      {"id": "Equation817", "kind": "abbrev", "module": "dep.equational_theories.Equations.Basic",
       "formal": {"type": "(G : Type u) → [a : Magma G] → Prop",
         "value": "fun G [Magma G] => ∀ (x : G), x = x ◇ ((x ◇ x) ◇ (x ◇ x))",
         "proof_valued": false, "projection": false, "private": false, "generated_kind": null}},
      {"id": "Magma", "kind": "class", "module": "dep.equational_theories.Magma",
       "formal": {"type": "Type u_1 → Type u_1",
         "value": null,
         "proof_valued": false, "projection": false, "private": false, "generated_kind": null}},
      {"id": "Magma.mk", "kind": "constructor", "module": "dep.equational_theories.Magma",
       "formal": {"type": "{α : Type u_1} → (α → α → α) → Magma α",
         "value": null,
         "proof_valued": false, "projection": false, "private": false, "generated_kind": null}},
      {"id": "Magma.op", "kind": "abbrev", "module": "dep.equational_theories.Magma",
       "formal": {"type": "{α : Type u_1} → [self : Magma α] → α → α → α",
         "value": "fun α [self : Magma α] => self.1",
         "proof_valued": false, "projection": true, "private": false, "generated_kind": null}}
    ],
    "external_nodes": [
      {"id": "And", "kind": "structure", "module": "Init.Prelude", "type": "Prop → Prop → Prop"},
      {"id": "Eq", "kind": "inductive", "module": "Init.Prelude", "type": "{α : Sort u_1} → α → α → Prop"},
      {"id": "Exists", "kind": "inductive", "module": "Init.Core",
       "type": "{α : Sort u} → (α → Prop) → Prop"},
      {"id": "Not", "kind": "def", "module": "Init.Prelude", "type": "Prop → Prop"}
    ]
  },
  "context": {
    "lean_toolchain": "leanprover/lean4:v4.34.1", "standalone_lean": false,
    "external_boundary": ["Mathlib", "Lean", "Std", "Init"]
  }
}
\end{lstlisting}
\end{promptbox}

\end{document}